\documentclass{article}

\PassOptionsToPackage{numbers, compress}{natbib}

\usepackage[preprint]{neurips_2022}

\usepackage[utf8]{inputenc} 
\usepackage[T1]{fontenc}    
\usepackage{hyperref}       
\usepackage{url}            
\usepackage{booktabs}       
\usepackage{amsfonts}       
\usepackage{nicefrac}       
\usepackage{microtype}      
\usepackage{xcolor}         

\usepackage{caption}
\usepackage{subcaption}
\usepackage{graphicx}
\usepackage{tablefootnote}
\usepackage{array}
\usepackage{float}
\graphicspath{ {./figs/} }

\usepackage{wrapfig}
\usepackage{amsmath}
\usepackage[capitalize,nameinlink]{cleveref}
\usepackage{enumitem}

\usepackage{graphicx,amsfonts,amscd,amssymb,bm,url,color,latexsym,bbm,amsthm,amsmath}
\usepackage{physics}
\allowdisplaybreaks

\renewcommand{\mathbf}{\boldsymbol}

\newcommand{\mb}{\mathbf}
\newcommand{\mc}{\mathcal}

\newcommand{\bb}{\mathbb}

\newcommand{\set}[1]{\left\{ #1 \right\}}

\newcommand{\paren}{\pqty}

\title{Imbalanced Classification in Medical Imaging \\ via Regrouping}

\author{%
  Le Peng$^1$, Yash Travadi$^2$, Rui Zhang$^3$, Ying Cui$^4$, Ju Sun$^1$  \\
  $^1$Computer Science \& Engineering, University of Minnesota, Twin Cities \\
  $^2$School of Statistics, University of Minnesota, Twin Cities \\
  $^3$Department of Surgery, University of Minnesota, Twin Cities\\
  $^4$Industrial and Systems Engineering, University of Minnesota, Twin Cities \\
  \texttt{\{peng0347,trava029,zhan1386,yingcui,jusun\}@umn.edu} \\
}

\begin{document}

\maketitle

\newcommand{\blue}[1]{\textcolor{blue}{#1}}
\definecolor{umn_maroon}{RGB}{122, 0, 25}

\begin{abstract}
We propose performing imbalanced classification by regrouping majority classes into small classes so that we turn the problem into balanced multiclass classification. This new idea is dramatically different from popular loss reweighting and class resampling methods. Our preliminary result on imbalanced medical image classification shows that this natural idea can substantially boost the classification performance as measured by average precision (approximately area-under-the-precision-recall-curve, or AUPRC), which is more appropriate for evaluating imbalanced classification than other metrics such as balanced accuracy. 
\end{abstract}

\paragraph{\textcolor{umn_maroon}{The shaky foundation of machine learning (ML) for medical imaging}} Modern data classification is founded on \textbf{accuracy} (ACC) maximization, or equivalently \textbf{error} minimization: 
\begin{align}
\min_{f \in \mc H} \quad \bb E_{(\mb x, \mb y) \sim \mc D} \mb 1\set{\mb y \ne f(\mb x)} = \sum_{i} \bb P(i = y) \bb E_{\mb x|y=i} \mb 1\set{\mb y \ne f(\mb x)} ,
\end{align} 
where $\mc H$ is the hypothesis class, $\mc D$ is the data distribution, and $\mb y$ is the one-hot-encoded vector of $y$. But accuracy can be misleading if the data are imbalanced across classes, e.g., $\mc D(\mb y)$ not uniform. This intrinsic class imbalance is prevalent in medical imaging classification (MIC), e.g., in binary cases, the negative rate is expected to far exceed the positive rate in most diseases, and in  multiclass cases, the prevalence rates of different diseases are generally disparate. A slightly refined notion is \textbf{balanced-accuracy} (BA), which leads to the \textbf{balanced-error} (BE) minimization: 
\begin{align}  \label{eq:bal_err}
\min_{f \in \mc H} \quad \frac{1}{|\set{y}|} \sum_{i} \bb E_{\mb x|y=i} \mb 1\set{\mb y \ne f(\mb x)}. 
\end{align} 
\begin{wrapfigure}{r}{0.4\textwidth}
    \vspace{-2em}
    \centering
    \includegraphics[width=0.95\linewidth]{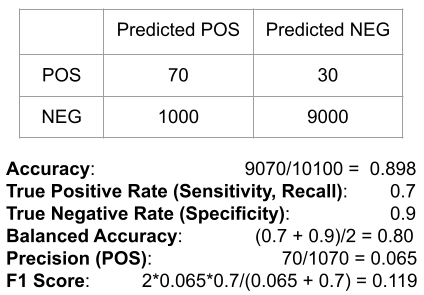}
    \caption{An example confusion table for binary classification, and the various associated performance metrics. }
    \label{fig:metrics_illus}
    \vspace{-1em}
\end{wrapfigure}
From a quick example in \cref{fig:metrics_illus}, it is clear that BA can be more responsive to recall performance than ACC, but it still fails to capture low precision which together with recall probably matters the most for medical diagnosis. Hence, for MIC, reporting precision-recall and the associated F1 score seems most appropriate~\cite{saito2015precision,williams2021effect}. But precision/recall and the F1 score depend on the decision threshold, for which our typical choice following the balanced scenarios is provably suboptimal with class imbalance~\cite{singh2021statistical,menon2013statistical}. Hence, the \textbf{area under the precision-recall curve} (AUPRC), which is often approximated by \textbf{average precision} (AP), seems a more sensible choice, as it only depends on the ordering of the prediction scores and not on individual thresholds~\cite{saito2015precision,williams2021effect}. In ML, optimizing AUROC is relatively easy and comes with mature computational tools, whereas optimizing AUPRC is still in its infancy~\cite{yang2022auc}. A further complication occurs for multiclass problems: optimizing multiple AUPRCs is intrinsically multi-objective optimization, and it is problematic to directly sum them up as the scaling of AUPRC is known to be sensitive to the class imbalance ratio. 

\paragraph{\textcolor{umn_maroon}{Fundamental ideas for imbalanced classification (IC) optimize BA}}
There are two classical ideas to counter the class imbalance: loss reweighting and class resampling, with a unified goal: (approximately) optimizing the BA. \textbf{(1) canonical loss reweighting}: the loss is reweighted by the inverse of class frequency, i.e., asymptotically 
\begin{align*}
\min_{f \in \mc H} \quad  \frac{1}{|\set{y}|}  \sum_{i} \textcolor{red}{1/\bb P(i = y)} \cdot \bb P(i = y) \bb E_{\mb x|y=i} \mb 1\set{\mb y \ne f(\mb x)} 
= \frac{1}{|\set{y}|} \sum_{i} \bb E_{\mb x|y=i} \mb 1\set{\mb y \ne f(\mb x)} ,
\end{align*}
which is equivalent to the BE minimization in \eqref{eq:bal_err}; 
\textbf{(2) canonical class resampling}: minority classes are upsampled by uniform sampling with replacement to match the sizes of the majority classes or vice versa. It is easy to verify that this is also equivalent to \eqref{eq:bal_err} asymptotically. \textbf{So if the goal is to maximize AUPRC, or even simply single-threshold precision-recall, these fundamental ideas are not aligned with the learning goal}. 

\paragraph{\textcolor{umn_maroon}{Recent developments in loss reweighting and class resampling}}
\textbf{(1)} Conceptually, loss reweighting assigns more weight to minority classes, so that misclassified minority samples induce more loss to ensure the classifier does not bias toward the majority classes~\citep{elkan2001foundations}. However, recent studies~\citep{byrd2019effect,xu2021understanding} find that the effect of reweighting vanishes as powerful deep neural networks (DNNs) gradually overfit the training data. More sophisticated reweighted losses focus on hard misclassified examples~\citep{lin2017focal} or maximizing classification margins~\citep{cao2019learning, sun2014imbalanced}, but often entail delicate hyperparameter tuning to achieve desirable results. \textbf{(2)} In resampling methods, random oversampling (ROS) minority classes is more effective and hence popular than downsampling majority classes~\cite{buda2018systematic}---throwing away data seems unwise for training data-hungry DNNs also. Advanced oversampling methods synthesize new samples from existing training data. Representative methods such as SMOTE~\citep{chawla2002smote}, balancing GAN~\citep{mariani2018bagan}, and VAE-based oversampling~\citep{zhang2018over} are competitive in performance. 

\paragraph{\textcolor{umn_maroon}{The boundary between imbalanced and long-tailed classification}} 
A closely related line of research is long-tail recognition in computer vision (CV)~\cite{zhang2021deep,yang2022survey}, which also concerns IC. However, the long-tail setting in CV is so extreme that the tail classes often contain only very few samples, effectively blending few-shot learning into IC and hence complicating the issue. By contrast, for real-world MIC, the dataset often contains a small number of classes with reasonable numbers of observations. Hence, for the rest of this paper, we focus on the classical imbalance regime: only a few target classes are considered, and despite the class imbalance each class still holds sufficiently representative samples. 
\begin{figure}[!htbp]
\centering
    \begin{subfigure}{.19\textwidth}
      \centering
      \includegraphics[width=\linewidth]{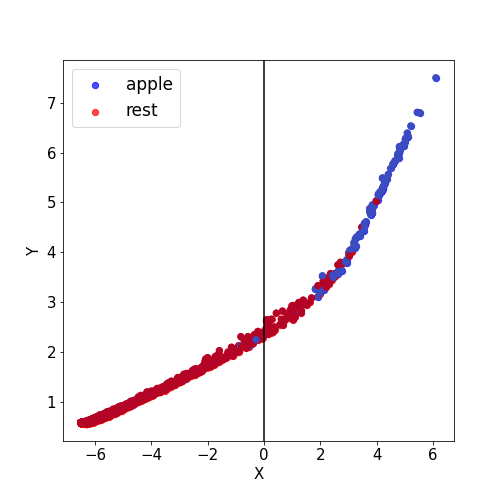}
      \caption{one-vs-rest}
    \end{subfigure}%
    \begin{subfigure}{.81\textwidth}
      \centering
      \includegraphics[width=\linewidth]{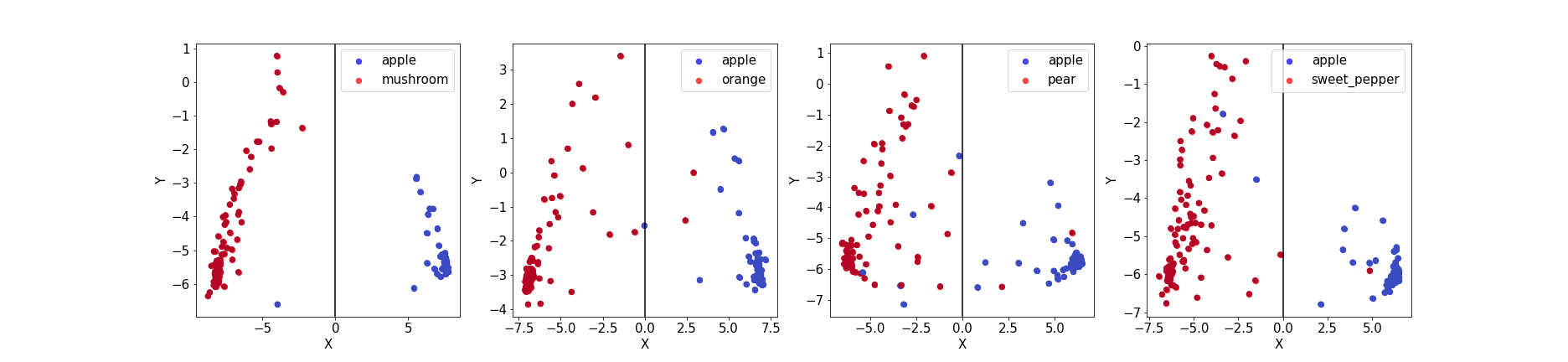}
      \caption{Multi-class Classification}
    \end{subfigure}
\caption{Classifying apple images vs the rest on CIFAR-100 test dataset. The features from the final convolutional layer of ResNet-34 are projected to 2D space. X-axis is along the direction normal to separating hyper-plane for 2 given classes and Y-axis is along the direction that is orthogonal to X-axis and has maximum deviation for the plotted points. \textbf{(a)} Model trained for binary classification of one-vs-rest directly: BA $= 0.946$, AP $=0.885$, \textbf{(b)} Model trained for 100-class classification used to classify one-vs-rest: BA $= 0.945$, AP $=0.967$. }
\label{fig:ill}
\vspace{-1em}
\end{figure}
\paragraph{\textcolor{umn_maroon}{Regrouping: the idea and intuitions}} 
In this paper, we introduce a simple yet effective regrouping method for IC. By splitting the samples from the majority classes into several groups using a clustering algorithm, our regrouping method can transform a hard IC problem into a familiar balanced one. The idea stems from two considerations: 1) \textbf{Intrinsic hierarchy of labels}:  in many cases imbalance is artifically introduced by the labeling process. For example, negative controls in medical data are typically the dominant class and are collected from multiple sources or demographic populations. If fine-grained labeled, the negative controls can be naturally regrouped into small clusters and hence induce more-balanced multiclass datasets;  2) \textbf{Efficient representation learning}: regrouping according to the intrinsic data structure helps DNN models to discover and learn the underlying patterns more efficiently. In Figure \ref{fig:ill} we show the positive effect of regrouping on the representations learned for the CIFAR-100 dataset. 


       

\paragraph{\textcolor{umn_maroon}{Regrouping: details}} Consider a binary classification problem with training data $\set{\paren{\mb x_i,y_i}}_{i=1}^n \sim_{iid} \mathcal{D}_{\mb x,y}$, where $\mb x_i\in\mathbb{R}^D$ and $y_i\in\left\{-1,+1\right\}$, and a DNN model $f_\theta:\mathbb{R}^D\rightarrow [0,1]$ parametrized by $\mb \theta$ and outputting a prediction score for the positive class. Assume the positive class is the minority class. 
Then the standard empirical risk minimization (ERM) can be formulated as:
\begin{align}
    \min_{\mb \theta} \frac{1}{n}\sum_{i:y_i=-1}\ell \paren{-1,f_{\mb \theta}(\mb x_i)} + \frac{1}{n}\sum_{i:y_i=+1}\ell \paren{+1,f_{\mb \theta}(\mb x_i)}. 
\end{align}
Obviously, when the imbalance ratio is high, the objective function is dominated by the negative class, and then the classifier is prone to prioritize the negative class over the positive. We propose to decompose the first term further by assigning pseudo-labels to the negative samples and performing multiclass classification so that the resulting classification problem is more balanced. To this end, we divide the negative samples into $K$ clusters, and assign samples belonging to each class a pseudo label $\tilde{y}_i$ from the set $\{2,3,\dots,K,K+1\}$, resulting in a classification problem with $(K+1)$ classes. The DNN model $f_{\mb \theta}$ is then adjusted to have $K+1$ output heads and trained: 
\begin{align}
    \min_{\mb \theta} \frac{1}{n} \sum_{i=1}^{n}\ell \paren{\tilde{y}_i,\tilde{f}_{\mb \theta}(\mb x_i)} = \frac{1}{n} \sum_{j=1}^{K+1}\sum_{i:y_i = j}\ell \paren{j,\tilde{f}_{\mb \theta}(\mb x_i)}, 
\end{align}
where $\tilde{f}_{\mb \theta}(\mb X)$ outputs the prediction scores for all pseudo-classes. At test time, the points classified to pseudo label $\hat{y}=1$ are classified as positive and the rest are classified as negative. A natural choice for the number of pseudo labels parameter $K$ is given by the imbalance ratio, i.e., $\#(\text{negatives})/\#(\text{positives})$. This algorithm generalizes naturally to multiclass cases by regrouping samples from multiple classes with relatively high frequencies.

\paragraph{\textcolor{umn_maroon}{Experiment}}
We compare our proposed regrouping (RG) method with the standard ERM (cross-entropy (CE) loss) and several state-of-the-art methods including: 1) Weighted cross-entropy loss (WCE), 2) Focal loss~\citep{lin2017focal}, 3) LDAM~\citep{cao2019learning}, and 4) AP-loss~\citep{chen2019towards} (AP), and 5) ROS. \textbf{(1) Dataset.} We train and evaluate our method on HAM10000~\citep{tschandl2018ham10000}\footnote{Available online: \url{https://dataverse.harvard.edu/dataset.xhtml?persistentId=doi:10.7910/DVN/DBW86T}} which is a large collection of multi-source dermatoscopic images of $7$ common skin lesions: Melanocytic nevi(nv), Melanoma(mel), Benign keratosis-like lesions(bkl), Basal cell carcinoma(bcc), Actinic keratoses(akiec), Vascular lesions(vasc), and Dermatofibroma(df). The dataset contains $10,015$ images with an imbalance ratio of up to $47:1$. \textbf{(2) Implementation details.}
Images are resized into $224\times224$ and normalized by the ImageNet mean and standard deviation. We apply mild random rotation and horizontal flips for data augmentation. For training, we use ResNet34 and SGD with a batch size of $256$. We initialize the learning rate as 0.01 and adjust it during training using the CosineAnnealingLR scheduler. We train the model for $300$ epochs and evaluate the final model by BA and class-wise AP. For the RG model, we apply $k$-means on the final features space using CLIP~\citep{radford2021learning}, and the number of clusters is determined by the ratio of the current class to the smallest class. 

\begin{table*}[!htpb]
\caption{\textbf{RG vs others on HAM10000 dataset}. We report both BA and class-wise AP scores. The $7$ classes are ordered from the largest class to the smallest class. $\uparrow$ means the larger the better. }
\label{tab: ham10000}
\begin{center}
\setlength{\extrarowheight}{-0.5mm}
\setlength{\tabcolsep}{1.0mm}{
\begin{tabular}{>{\small}l | >{\small}c  >{\small}c >{\small}c >{\small}c >{\small}c >{\small}c >{\small}c >{\small}c}
\midrule
\bottomrule
\vspace{-8pt}
\\
\small{\textbf{Method}}
&\small{\textbf{BA ($\%$) $\uparrow$}}
&{}
&{}
&{}
&\small{\textbf{AP ($\%$) $\uparrow$}}
&{}
&{}
&{}
\\
\cline{3-9}
&{}
&{nv}
&{mel}
&{bkl}
&{bcc}
&{bakiec}
&{vasc}
&{df}
\\
\toprule
\small{ERM(CE)}
&{52.0}
&{96.0}
&{56.1}
&{62.8}
&{69.2}
&{52.8}
&{70.0}
&{33.2}
\\
\small{WCE}
&\textcolor{red}{63.8}
&{96.0}
&{42.5}
&{49.6}
&{55.7}
&{51.3}
&{80.8}
&{43.4}
\\
\small{Focal}
&{55.2}
&{97.0}
&{58.3}
&{63.5}
&{73.2}
&{50.6}
&{72.0}
&{42.8}
\\
\small{LDAM}
&{56.8}
&{95.7}
&{56.6}
&{58.5}
&{69.8}
&{52.2}
&\textcolor{red}{84.1}
&{35.4}
\\
\small{AP}
&{52.3}
&{95.8}
&{58.3}
&{62.3}
&{65.3}
&{58.3}
&{82.9}
&{11.1}
\\
\small{ROS}
&{61.3}
&{96.7}
&{56.1}
&{61.9}
&{71.1}
&{54.1}
&{82.7}
&\textcolor{red}{51.4}

\\

\midrule
\small{RG+CE(Ours)}
&{63.2}
&\textcolor{red}{97.2}
&\textcolor{red}{64.3}
&\textcolor{red}{70.9}
&\textcolor{red}{78.8}
&\textcolor{red}{64.1}
&{82.4}
&{43.8}

\\
\bottomrule
\end{tabular}
}
\end{center}
\end{table*}

\textbf{Results} \cref{tab: ham10000} summarizes our results. We observe that: 1) \textbf{naive ERM produces biased results}. EMR performs similarly to other methods in large classes (e.g., nv), but it is worse in small classes (e.g., vasc and df). 2) \textbf{BA may give misleading results}. WCE achieves the best BA score, but the class-wise AP's can be much worse than methods like ROS and our RG+CE. 3) \textbf{RG+CE shows its superiority in AP}. Our proposed method improves the APs for minority classes by over $10\%$ compared to EMR and produces a competitive BA score. Notably, the AP method designed to optimize AP directly does not perform well. We believe it is due to their optimization artifact when approximating the AP with a surrogate objective.


   


{\small
\bibliography{ref}
\bibliographystyle{unsrt}
}

\end{document}